\documentclass[10pt,twocolumn,letterpaper]{article}
\usepackage{graphicx}
\usepackage[dvipsnames,table,xcdraw]{xcolor}

\usepackage{iccv}

\makeatletter
\@namedef{ver@everyshi.sty}{}
\makeatother
\usepackage{tikz}
\usepackage{amsmath,amssymb} 

\usepackage{caption}
\usepackage{subcaption}

\usepackage{times}
\usepackage{epsfig}
\usepackage{dsfont}

\usepackage{url}
\usepackage{booktabs}
\usepackage{amsfonts}
\usepackage{nicefrac}
\usepackage{microtype}
\usepackage{caption}
\usepackage{comment}
\usepackage{multirow}
\usepackage{multicol}
\usepackage{stfloats}

\usepackage[pagebackref=true,breaklinks=true,letterpaper=true,colorlinks,bookmarks=false]{hyperref}

\usepackage[capitalize]{cleveref}
\crefname{section}{Sec.}{Secs.}
\Crefname{section}{Section}{Sections}
\Crefname{table}{Table}{Tables}
\crefname{table}{Tab.}{Tabs.}

\newcommand{\midsepremove}{\aboverulesep = 0mm \belowrulesep = 0mm}
\midsepremove
\newcommand{\midsepdefault}{\aboverulesep = 0.605mm \belowrulesep = 0.984mm}
\midsepdefault

\iccvfinalcopy 

\def\iccvPaperID{6146} 

\begin{document}

\newcommand{\OURS}{Mesh2Tex}
\newcommand{\TODO}[1]{{\textcolor{red}{TODO: #1}}}
\newcommand{\Alex}[1]{{\textcolor{blue}{Alex: #1}}}
\newcommand{\shubham}[1]{\textcolor{Blue}{ST\@: {#1}}}
\newcommand{\ANGIE}[1]{{\emph{\textcolor{ForestGreen}{Angie: #1}}}}

\title{\OURS{}: Generating Mesh Textures from Image Queries}

\author{Alexey Bokhovkin\\
Technical University of Munich\\
{\tt\small aleksei.bokhovkin@tum.de}
\and
Shubham Tulsiani\\
Carnegie Mellon University\\
{\tt\small shubhtuls@cmu.edu}
\and
Angela Dai\\
Technical University of Munich\\
{\tt\small angela.dai@tum.de}
}

 \twocolumn[{%
 	\renewcommand\twocolumn[1][]{#1}%
 	\maketitle
 	\begin{center}
 		\vspace{-0.8cm}
 		\includegraphics[width=0.91\linewidth]{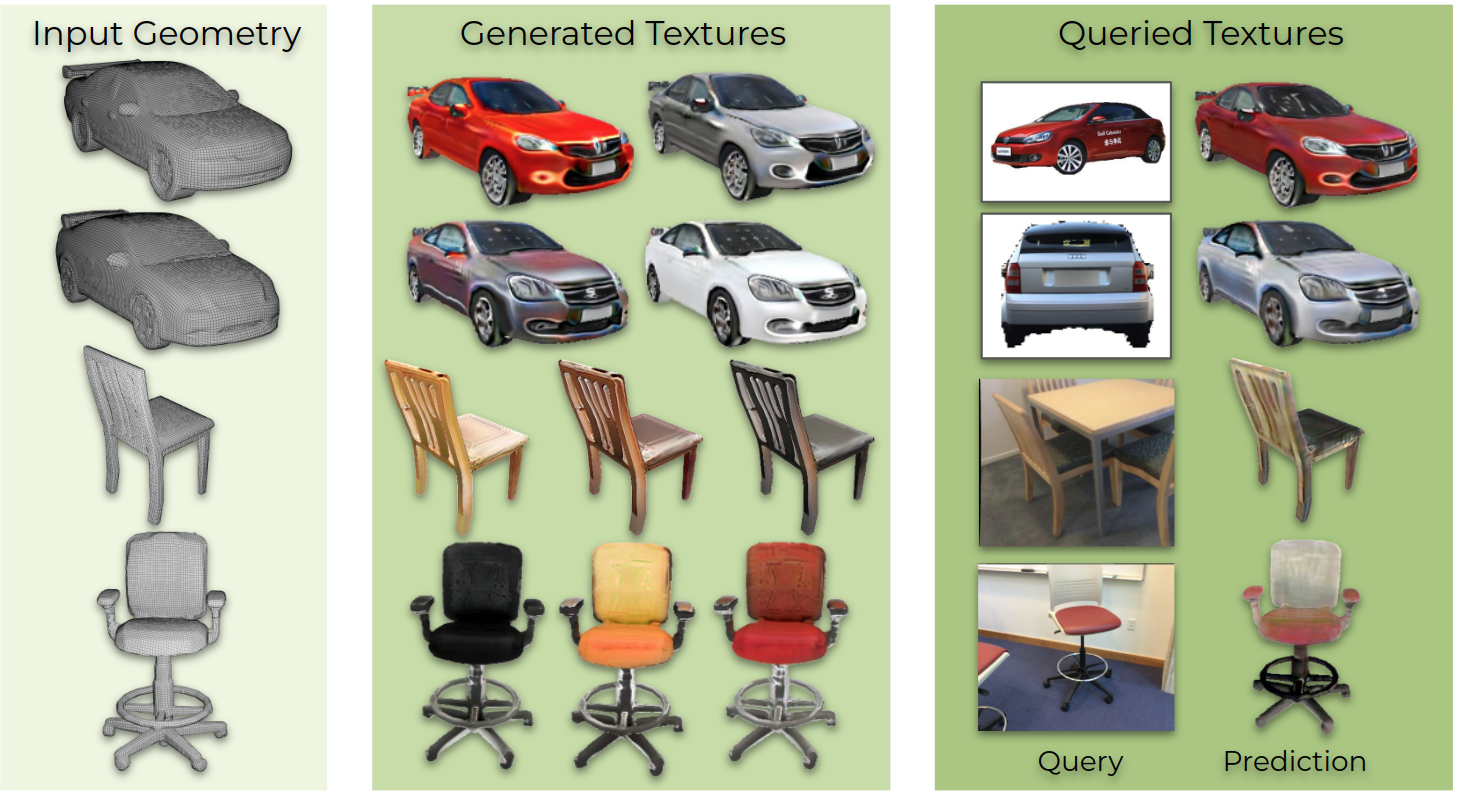}
 		\vspace{-0.2cm}
 		\captionof{figure}{
             \OURS{} learns realistic object texturing on a shape geometry through a hybrid mesh-field texture representation supporting high-resolution texture generation on various shape meshes.
             Furthermore, our learned texture manifold supports texture transfer optimization from image queries without requiring any matching geometry or pose alignment to the image, producing perceptually consistent texturing in this challenging content creation scenario.
 		}
             \vspace{-0.1cm}
 		\label{fig:teaser}
 	\end{center}
 }]

\maketitle
\ificcvfinal\thispagestyle{empty}\fi

\begin{abstract}

Remarkable advances have been achieved recently in learning neural representations that characterize object geometry, while generating textured objects suitable for downstream applications and 3D rendering remains at an early stage. 
In particular, reconstructing textured geometry from images of real objects is a significant challenge -- reconstructed geometry is often inexact, making realistic texturing a significant challenge.
We present \OURS{}, which learns a realistic object texture manifold from uncorrelated collections of 3D object geometry and photorealistic RGB images, by leveraging a hybrid mesh-neural-field texture representation.
Our texture representation enables compact encoding of high-resolution textures as a neural field in the barycentric coordinate system of the mesh faces.
The learned texture manifold enables effective navigation to generate an object texture for a given 3D object geometry that matches to an input RGB image, which maintains robustness even under challenging real-world scenarios where the mesh geometry approximates an inexact match to the underlying geometry in the RGB image.
\OURS{} can effectively generate realistic object textures for an object mesh to match real images observations towards digitization of real environments, significantly improving over previous state of the art.

{\normalfont Project page:} {\normalfont \url{ alexeybokhovkin.github.io/mesh2tex/}}
\end{abstract}

\section{Introduction}
The ability to obtain 3D representations of real-world objects lies at the core of many important applications in graphics, robotics, movies and video games, and mixed reality. Approaches tackling the tasks of 3D generation \cite{park2019deepsdf, luo2021diffusion, xie2020genpointnet, autosdf2022, nash2020polygen, deng2021deformed} or single-view reconstruction \cite{chen2018implicit_decoder, mescheder2019occupancy, NIPS2019_8340, Kuo2021Patch2CADPE, dahnert2021panoptic} enable users to easily create digital 3D assets, either from scratch or conditioned on an image. However, these approaches primarily focus on generating/inferring the geometry of objects, and this alone is insufficient to capture realistic environments, which requires high-quality texturing. 

In this work, we propose \OURS{} tackle the complementary task of high-fidelity texture generation given known object geometry. In addition to allowing texturing generation for a given object mesh, \OURS{} also enables image-based texture synthesis -- synthesizing textures that perceptually match a single RGB image observation. In contrast to existing appearance modeling works that characterize texture as a field over the volume of 3D space \cite{chan2021eg3d, saito2019pifu, saito2020pifuhd, oechsle2019texture}, we observe that object textures lie on object surfaces, and instead generate textures only on the mesh surface, with a hybrid explicit-implicit texture representation that ties the texture field to the barycentric coordinates of the mesh faces. This produces textured meshes directly compatible with downstream applications and 3D rendering engines.

As large quantities of high-quality textured 3D objects are very expensive to obtain, requiring countless hours of skilled artists' work, we train our texture generator with uncorrelated collections of 3D object geometry and photorealistic 2D images, leveraging differentiable rendering to optimize mesh textures at arbitrary resolutions to match the quality of real 2D images in an adversarial fashion. 
That is, we generate coarse features and colors for each mesh face, which are then refined to high-resolution textures through a neural field defined on the barycentric coordinates of each mesh face. 
This hybrid texture representation enables differentiable rendering for texturing on explicit mesh surfaces while exploiting the efficiency of  neural field representations.

Our learned texture generator can then be used to produce realistic textures to match an input RGB observation of an object, by optimizing over our learned texture manifold conditioned on the mesh geometry, to find a latent texturing that perceptually matches the RGB image. 
As state-of-the-art object geometry reconstructions are typically inexact (e.g., noise, oversmoothing, retrieval from a database), our learned texture manifold enables effective regularization over plausible textures which effectively capture the perceptual input while providing realistic texturing over the full object. 
We formulate a patch-based style loss to capture perceptual similarities between our optimized texture and the RGB image, guided by dense correspondence prediction to correlate them.
Experiments demonstrate that we outperform state of the art in both unconditional texture generation as well as image-conditioned texture generation.

In summary, our contributions are:
\begin{itemize}
\item a new hybrid mesh-neural-field texture representation that enables diverse, realistic texture generation on object mesh geometry by tying a neural texture field to the barycentric coordinate system of the mesh faces. This enables learning a rich texture manifold from collections of uncorrelated real images and object meshes through adversarial differentiable rendering.
\item our learned texture manifold enables effective inference-time optimization to texture an object mesh to perceptually match a single real-world RGB image; crucially, we maintain robustness to real-world scenarios of differing views and even geometry with a patch-based perceptual optimization guided by dense correspondences.
\end{itemize}


\section{Related Work}

\paragraph{Optimization-based Texturing.}
Various approaches have been proposed to texture 3D shapes by optimizing from aligned image views of the shape. 
The traditional optimization approach of Zhou et. al.~\cite{zhou2014colormap} solves a global optimization for a mapping from observed RGB images onto reconstructed geometry.
More recently, Huang et. al.~\cite{huang2020adversarial} introduce a modern adversarial discriminator loss, coupled with differentiable rendering, to optimize for texture on reconstructed geometry to produce realistic texturing robust to camera pose misalignments in real-world RGB-D capture scenarios.
While these methods can produce impressive texturing results, they require aligned 2D/3D data and cannot produce textures in unobserved regions.

\paragraph{Learned Texture Completion and Generation.}
Learning-based methods have also explored generative texturing, aiming to learn generalized features for unconditional synthesis or conditional inpainting.
Photoshape~\cite{photoshape2018} proposes to learn material retrieval to apply materials to 3D shapes.
Recently, several methods have been developed to generate textures on various representations: SPSG~\cite{dai2021spsg} produces per-voxel colors with an adversarial differentiable rendering formulation, Pavollo et. al.~\cite{pavllo2021textured3dgan} learn texturing in the UV domain, and Henderson et. al.~\cite{henderson2020leveraging} and Texturify~\cite{siddiqui2022texturify} propose to generate per-face mesh colors with variational and adversarial training, respectively.
The recent success of neural implicit field representations has also inspired implicit texture field models \cite{OechsleICCV2019,chan2021eg3d,gao2022get3d}.
Our approach aims to model texture on the surface of a mesh, coupled with the representation capacity of a shared implicit field per mesh face.

\paragraph{Query-based 3D Understanding.}
The recent success of vision-language models such as CLIP~\cite{radford2021learning} has sparked interest in query-guided 3D understanding from text and images.
Text and image guidance have been exploited for knowledge distillation for 3D semantic understanding \cite{koo2022partglot,rozenberszki2022language,jatavallabhula2023conceptfusion}.
Various shape generation tasks have also been formulated with text queries, by leveraging CLIP  as supervision \cite{sanghi2021clip,jain2021dreamfields, Wang2021CLIPNeRFTD}.
In particular, Text2Mesh~\cite{text2mesh} leverages CLIP guidance for text-based texture optimization on 3D shapes.
We formulate \OURS{} to support image-based queries for texture transfer optimization to various 3D shapes, to reflect potential practical texturing scenarios where images are largely easy to capture or find, with high visual specificity.
We thus tackle challenges in geometric and pose misalignments from various potential image queries.

\section{Method}

\subsection{Overview}

\begin{figure*}
\begin{center}
    \includegraphics[width=1.0\textwidth]{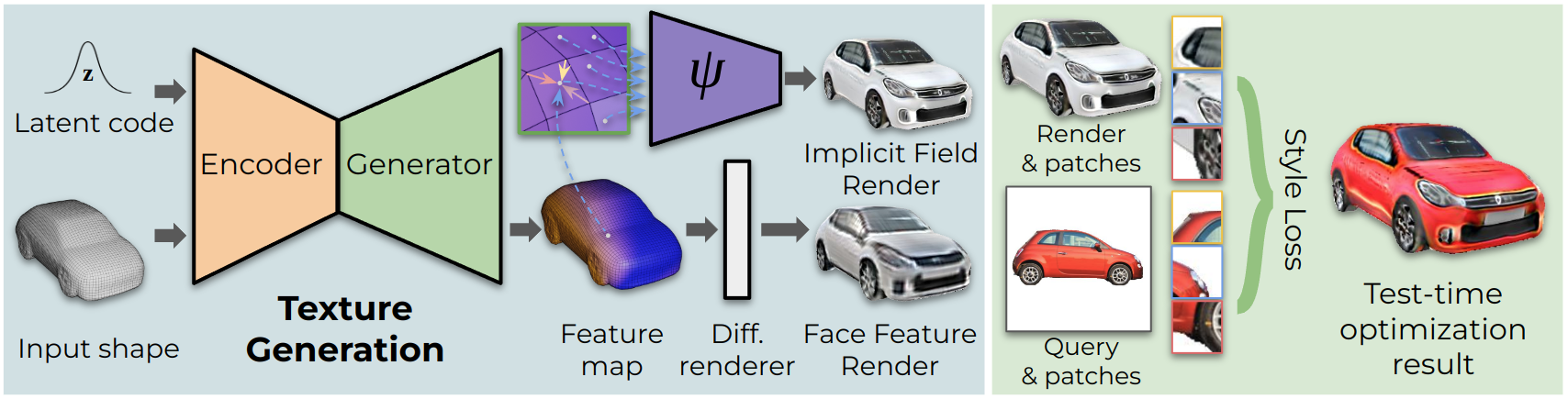}
    \vspace{-0.7cm}
    \caption{Method overview.
    Left: we first learn a texture generator for textures represented as a shared local neural field in the barycentric coordinate system of each mesh face.
    This enables the capture of high-resolution texture detail, while remaining tied to surface geometry, avoiding ambiguities of volumetric field representations.
    Our learned texture manifold can then be used for texture transfer from a single RGB query image, based on a correspondence-guided patch-based style loss to produce perceptually consistent texturing even for shapes with differing geometry and unknown image pose.
    \vspace{-0.6cm}
    }
    \label{fig:overview}
\end{center}%
\end{figure*}

We aim to learn a high-quality texture manifold conditioned on a given shape mesh geometry, which supports unconditional texture generation as well as optimization within the texture manifold to fit to image queries for practical texturing scenarios.
An overview of our approach is visualized in Figure~\ref{fig:overview}.
We propose to learn this texture manifold on a hybrid representation that combines a mesh representation with a neural field operating on the barycentric coordinates of each mesh face.
Our texture generator then generates initial coarse features per-face for a given shape mesh, which are then refined by a shared neural field across the mesh faces.
This enables high-resolution texture generation by sampling the neural field on the mesh surface.
In the absence of a large-scale dataset of high-quality 3D textured shapes, we supervise the texture generation with an adversarial loss with photorealistic 2D images, through differentiable rendering.

We can then traverse our learned texture manifold to generate textures for a given shape geometry to perceptually match to image queries.
Since an RGB image query may be taken at an arbitrary camera pose, we estimate the object pose in the RGB image along with its normalized object coordinates~\cite{wang2019normalized}  to guide the texture optimization.
Furthermore, we must also support texture transfer scenarios, as having an exact geometric shape match for an arbitrary RGB image query cannot be assumed in practice.
We thus employ both a global and local patch-based style loss to generate a realistic texture that perceptually matches  the query image.

\subsection{Mesh-Field Texture Generation}

\begin{figure}
\begin{center}
    \includegraphics[width=0.99\linewidth]{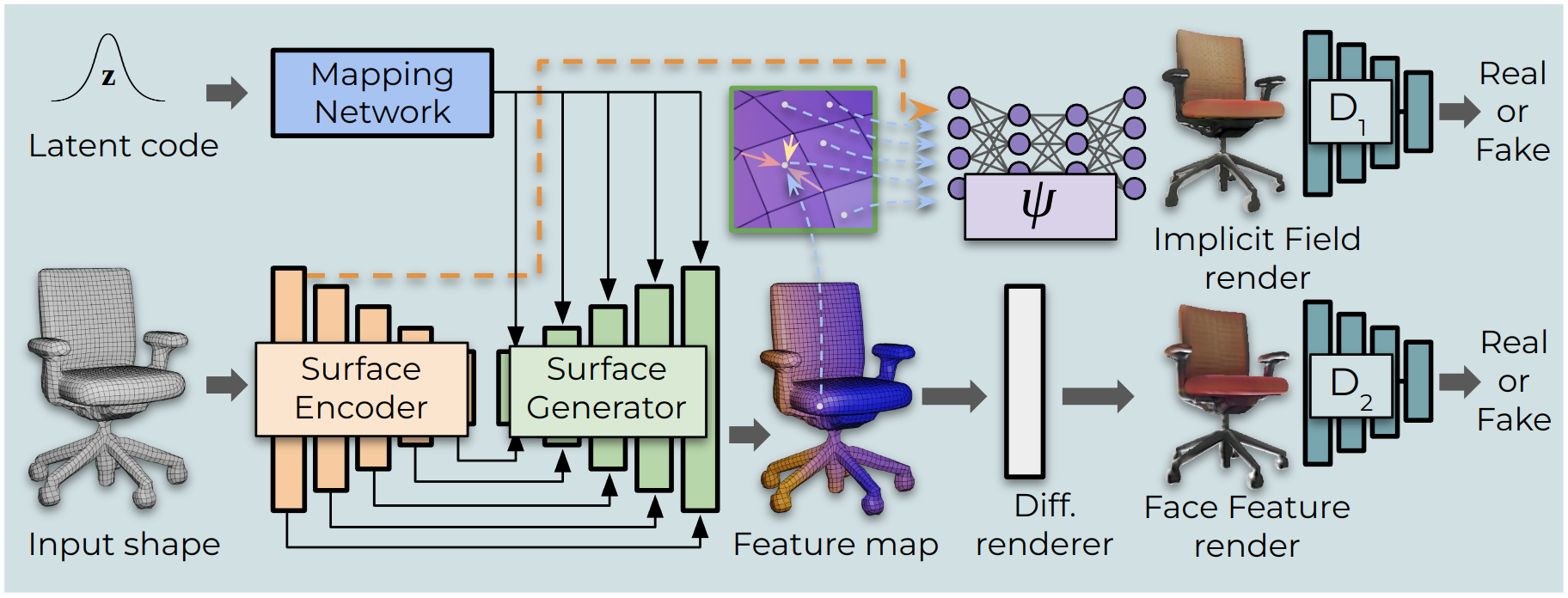}
    \vspace{-0.1cm}
    \caption{Mesh-field texture generation. We employ face convolutions following \cite{siddiqui2022texturify} for initial coarse feature generation, followed by a neural field that interprets the per-face features to produce refined color outputs at arbitrary locations on the mesh surface. 
    The hybrid mesh-field texture generation is supervised in an adversarial fashion through differentiable rendering,  with photorealistic 2D images representing the real distribution.
    \vspace{-0.5cm}
    }
    \label{fig:training}
\end{center}%
\end{figure}

We learn a texture manifold on a mesh-field representation. 
Given a shape mesh $\mathcal{M}$, we use a latent-variable conditioned encoder-decoder to obtain per-face features $F$. We then employ a shared neural field $\psi$ that operates on each face to produce arbitrary-resolution texturing. High-quality texture generation is then learned by supervision with photorealistic 2D images in an adversarial fashion through differentiable rendering.
Our texture generator training is shown in Figure~\ref{fig:training}.

A shape mesh $\mathcal{M}$ is represented as a quadrilateral mesh; inspired by the hierarchical mesh representation of Texturify~\cite{siddiqui2022texturify}, we also employ QuadriFlow~\cite{huang2018quadriflow} to parameterize shape meshes as hierarchies of 4-way rotationally symmetric (4-RoSy) fields.
We can then leverage convolutional and pooling operators on the neighborhoods of each quad face, enabling encoding and decoding features on such mesh faces with arbitrary topologies.
We employ an encoder-decoder face convolutional network $G$ on $\mathcal{M}$, which takes as input geometric quad face features (per-face normals, fundamental forms, and curvature), and predicts per-face features $F_c$.
Similar to StyleGAN~\cite{Karras2020AnalyzingAI} we incorporate a latent texture code $z$ into the decoder through a mapping network.

As we aim to generate high-quality textures not limited to the mesh face resolution, we leverage a neural field $\psi$. $\psi$ takes as input a barycentric coordinate $p$ on a mesh face, along with mean face feature $F_v$ of all incident quad faces: 
\begin{equation}
\begin{split}
\psi(p, b_1 F_{v_1} + b_2 F_{v_2} + b_3 F_{v_3}) = {} & c \in \mathbf{R}^{3}_{[-1, 1]},
\end{split}
\end{equation}
where $\mathcal{A} = \{v_1, v_2, v_3\}$ is the triangle half of a quad face $\{v_1, v_2, v_4, v_3\}$ in which $p$ lies, $\{b_1, b_2, b_3\}$ are the barycentric coordinates of $p$ with respect to triangle $\mathcal{A}$, and $c$ is the final output color.
We can then generate arbitrary-resolution textures by densely sampling $\psi$ on each mesh face. 

To supervise the learning of this mesh-field texture manifold, we differentiably render our generated textures for an adversarial loss with photorealistic 2D images.
To render the generated textures, we sample $\psi$ at the locations corresponding to each pixel in the rendered view. In addition to implicit field $\psi$ rendering, we also pass the output feature map $F_c$ into one convolutional layer and render it using PyTorch3D~\cite{ravi2020pytorch3d} differentiable renderer and supervise it similarly as a proxy loss.
We then train the texture manifold in an end-to-end fashion using a non-saturating GAN loss with
gradient penalty and path length regularization.

\subsection{Texture Transfer from a Single Image}

\begin{figure}
\begin{center}
    \includegraphics[width=0.97\linewidth]{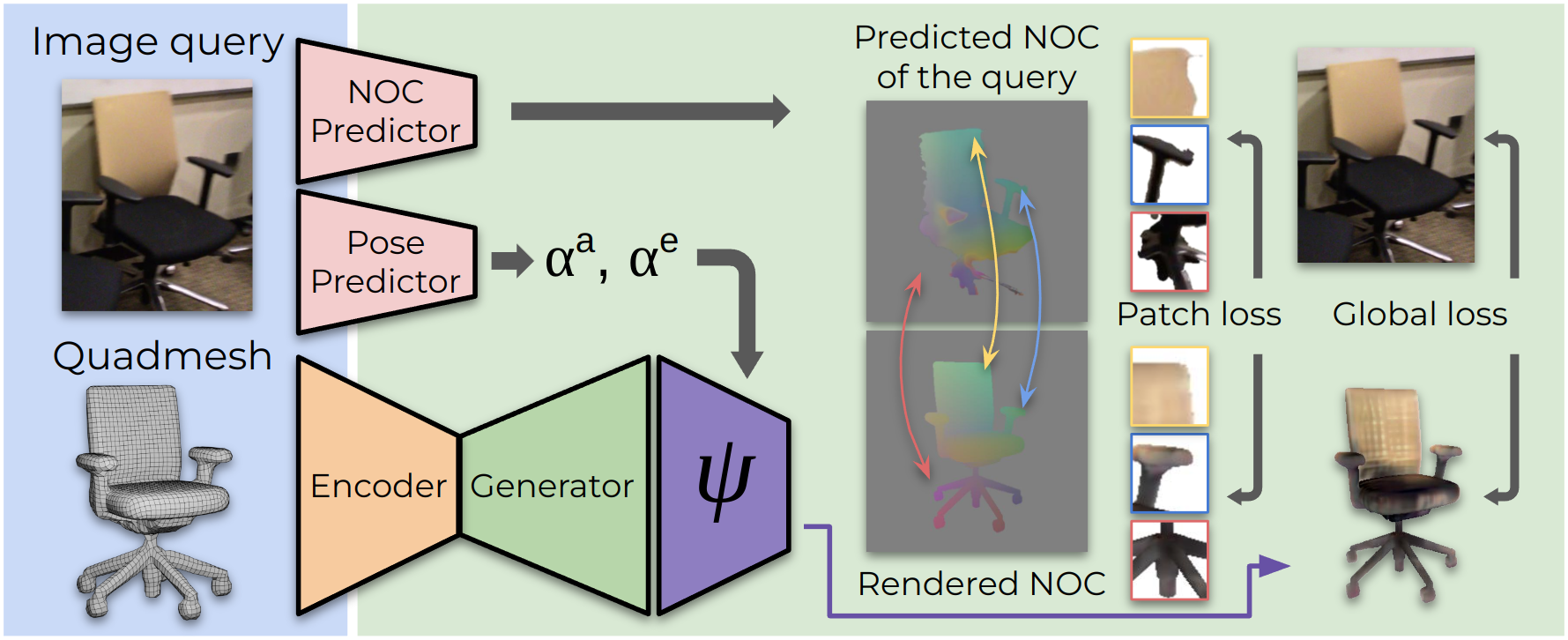}
    \vspace{-0.1cm}
    \caption{Texture transfer optimization from a single image.
    We traverse our learned texture manifold for texture transfer from a single RGB image query.
    Since image alignment may be unknown, we estimate a coarse pose along with finer-grained dense NOC correspondences. 
    We leverage the NOCs to guide patch sampling for our patch-based style loss to produce texturing that perceptually matches to the query images, without requiring any exact geometric shape matches to the image.
    \vspace{-0.5cm}
    }
    \label{fig:optimization}
\end{center}%
\end{figure}

Once learned, we can not only use our texture manifold for unconditional generation, but also importantly traverse through the manifold to produce shape texturing that matches to a query RGB image $I$.
This represents real-world texturing scenarios where a user may wish to texture a shape based on an easy-to-capture image inspiration.
Here, notable challenges are lack of knowledge of the ground truth object pose in $I$, and handling inexact geometric matches between the shape mesh to be textured and the object in $I$ -- as it is not practical, nor always desirable, to assume the ability to reconstruct or retrieve the exact geometry of the object in an arbitrary image.
We thus aim to produce textures on a shape to perceptually match an image $I$ in these challenging scenarios.
Our texture transfer optimization is illustrated in Figure~\ref{fig:optimization}.

From $I$, we first estimate the object pose as the azimuth and elevation angles, $\alpha^{a}$ and $\alpha^{e}$, respectively. 
To this end, we use a ResNet-18~\cite{he2016deep} network on $I$, pre-trained on synthetic rendered shapes to classify $\alpha^{a}$ and $\alpha^{e}$ into 12 and 5 bins for a coarse rotation estimate.
For finer-grained reasoning, we predict normalized object coordinates~\cite{wang2019normalized} $I_\textrm{NOC}$, dense correspondences from the object pixels in $I$ to the canonical space of the object.
For NOC prediction, we use a UNet-style network with an EfficientNet-b4~\cite{Tan2019EfficientNetRM} backbone, trained on synthetic rendered shape data.

We then formulate our texture transfer optimization.
We differentiably render our generated texture from initial pose estimate $\alpha^{a}$ and $\alpha^{e}$ to produce generated image $X$.
Since we optimize for perceptual similarity, we employ a style loss~\cite{gatys2016image} in a global and local fashion to compare $X$ and $I$.

We then optimize for latent texture code $z$ with the loss:
\begin{equation}
\begin{split}
L = {} & w_{glob}\sum_{h=1}^{N_h}\sum_{l=1}^{N_l}L_{glob}(I^{h,l}, X^{h,l}) \\
 & + w_{patch}\sum_{p=1}^{N_p}\sum_{h=1}^{N_h}\sum_{l=1}^{N_l}L_{patch}(I_{p}^{h,l}(x, y), X_{p}^{h,l}(x', y')),
\end{split}
\label{eq:ttloss}
\end{equation}
where $L_{glob}$ denotes a global style loss, $L_{patch}$ a local patch-based style loss, and $w_{glob}$ and $w_{patch}$ constants to balance the loss values.

\paragraph{Global style loss.} 
$L_{glob}$ considers the full images $I$ and $X$ at $N_h=3$ resolution levels (original, half, and quarter resolution): $I^h$ and $X^h, h\in[1,2,3]$.
The global style loss is then computed on each pair $(X^h, I^h)$ on $N_l=5$ VGG feature layers.

\paragraph{Patch style losses.}
Since the precise structure of the object in $I$ may not match with that of $X$ (e.g. geometry mismatch or inconsistent pose prediction), we further employ local style losses on image patches.
We extract $N_p = 2$ patches $I_p(x, y)$ from $I$ randomly per iteration, located at patch centers $(x, y)$ within the object. 
We then use the corresponding NOC estimate at the patch location, $I_\textrm{NOC}(x,y)$, and find a corresponding patch $X_p(x',y')$ where $(x',y')$ is the pixel location whose shape NOC is closest to $I_\textrm{NOC}(x,y)$ w.r.t. $\ell_2$ distance.
Note that since $X$ has been rendered from a shape mesh, we can use ground truth shape NOCs for $X$, and only require predicted NOCs for $I$.
This provides a coarse correspondence between patches to guide the style loss.
Similar to $L_{glob}$, we apply $L_{patch}$ at $N_h=3$ resolution levels and with $N_l=5$ VGG feature layers.

\paragraph{Face feature refinement.}
With texture code $z$ optimized from Equation~\ref{eq:ttloss}, we further allow for texture refinement by unfreezing the last two face convolutional layers in the texture generator.
We optimize for refined weights in the surface features using only $L_{patch}$, which enables better capturing of local detail in $I$.

\paragraph{Implementation Details}
Our \OURS{} model is implemented using PyTorch and trained using an Adam~\cite{Kingma2014AdamAM} optimizer with learning rates of 1e-4, 12e-4, 1e-4, 14e-4 for the encoder, generator, neural field $\psi$ parameters and both discriminators, respectively. \OURS{} is trained on 2 NVIDIA A6000s for 50k iterations ($\sim$ 100 hours) until convergence.

At test-time, we optimize latent codes for 100 iterations and refined weights for additional 300 iterations, which takes $\sim$ 400s in total. During texture optimization, we extract 64 $\times$ 64 patches. We provide further details in the supplemental.
\section{Results}

We evaluate the texture generation capabilities of \OURS{} on unconditional generation and image-based texture generation, on both synthetic and real data.
For both scenarios, our texture generation is trained from real images.

\paragraph{Datasets.}
For evaluation, we use object geometry from ShapeNet~\cite{shapenet2015} for texturing, and real-world query images from ScanNet~\cite{dai2017scannet} for chairs, and from CompCars~\cite{yang2015large} for cars.
For synthetic experiments requiring exact geometric matches, we use ShapeNet textured objects and render query image views.
For real image queries with close matching geometry, we use Scan2CAD~\cite{avetisyan2019scan2cad} annotations for chair meshes to ScanNet images.
For CompCars, we use the coarse pose information (front, back, left, right, etc.) to estimate close image alignments.

Note that all methods, except GET3D, were trained with the same set of ShapeNet meshes, and images from PhotoShape~\cite{photoshape2018} and CompCars~\cite{yang2015large} for chairs and cars. GET3D requires a much denser sampling of images per shape rather than a single view per shape, so we instead compare with the authors' released pre-trained GET3D model.

\paragraph{Evaluation metrics.} We evaluate the perceptual quality of our generated textures with several metrics. 
To measure the realistic quality of the textures, we compare rendered views of generated textures on various shapes to real-world images, using the standard FID~\cite{heusel2017gans} and KID~\cite{binkowski2018demystifying} scores used for assessing generated image quality.
For FID and KID evaluation, we use real images from PhotoShape~\cite{photoshape2018} and CompCars~\cite{yang2015large} for chairs and cars, respectively, and 6 rendered views from each synthesized textured shape.
Additionally, for texture transfer from a single image, we compute CLIP~\cite{radford2021learning} similarity as cosine distances between the query image and rendered texture.
In synthetic experiment setups where exact geometric matches are available, we further compute an LPIPS~\cite{zhang2018perceptual} perceptual metric between synthesized views and the query image.

\paragraph{Baselines.}
We compare with several state-of-the-art texturing methods leveraging various texture representations: 
Yu et. al.~\cite{yu2021learning} learns texture generation on a UV map parameterization, EG3D~\cite{chan2021eg3d} leverages an efficient triplane representation for view synthesis, Texturify~\cite{siddiqui2022texturify} generates per-face colors on a mesh, and GET3D~\cite{gao2022get3d} jointly estimates color and geometry with triplane representations.

\subsection{Unconditional Texture Generation}

\begin{figure}
\begin{center}
    \includegraphics[width=0.98\linewidth]{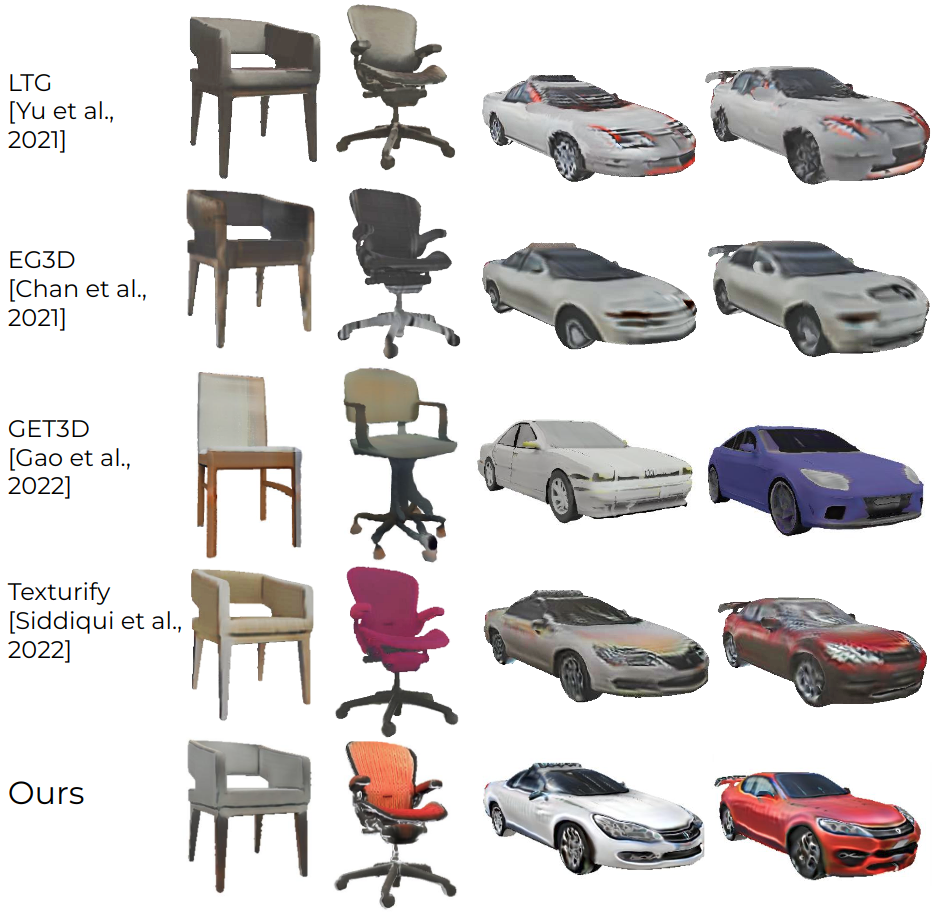}
    \vspace{-0.4cm}
    \caption{
    Unconditional texturing for meshes from ShapeNet~\cite{shapenet2015}, in comparison with state of the art.
    Our approach generates more realistic, detailed textures.
    \vspace{-0.7cm}
    }
    \label{fig:generation}
\end{center}%
\end{figure}

Table~\ref{tab:generation} and Figure~\ref{fig:generation} show quantitative and qualitative  comparisons of unconditional texture generation to state of the art. 
Our hybrid mesh-field texture representation enables richer texture generation with finer-scale details than state-of-the-art baselines.

\midsepremove
\begin{table}[tp]
\centering
\resizebox{0.48\textwidth}{!}{
\begin{tabular}{l|l|cc|cc}
\toprule
    \multirow{2}{*}{Method} & \multirow{2}{*}{Parameterization} & \multicolumn{2}{c|}{Chairs} & \multicolumn{2}{c}{Cars} \\
    & & FID & KID & FID & KID \\
\midrule
    EG3D & Tri-plane Implicit & 36.45 & 2.15 & 83.11 & 5.95 \\
    Yu et al. & UV & 38.98 & 2.46 & 73.63 & 5.77 \\
    GET3D & Tri-plane Implicit & 46.20 & 3.02 & 89.62 & 6.92 \\
    Texturify & 4-RoSy Field & 31.08 & 1.75 & 59.55 & 4.97 \\
\midrule
    {\bf Ours} & 4-RoSy Implicit Field & {\bf 30.01} & {\bf 1.69} & {\bf 41.35} & {\bf 3.41} \\
\bottomrule
\end{tabular}
}
\vspace{-0.2cm}
\caption{Unconditional texture generation for 3D shapes. Our mesh-field approach outperforms state of the art in generation quality. The KID values are scaled by $10^2$.
\vspace{-0.4cm}
}
\label{tab:generation}
\end{table}

\subsection{Texture Transfer from a Single Image}

We further evaluate texturing of objects from image queries, which opens up new possibilities in the content creation process.
We consider both synthetic image queries, which enable texturing of exact geometric matches, as well as real image queries, where no object geometry matches exactly to the queries. 
For all baselines, we apply a similar texture optimization as our approach, leveraging global and patch-style losses without NOC guidance.

\subsubsection{Synthetic Image Query Experiments}

\begin{figure}
\begin{center}
    \includegraphics[width=0.98\linewidth]{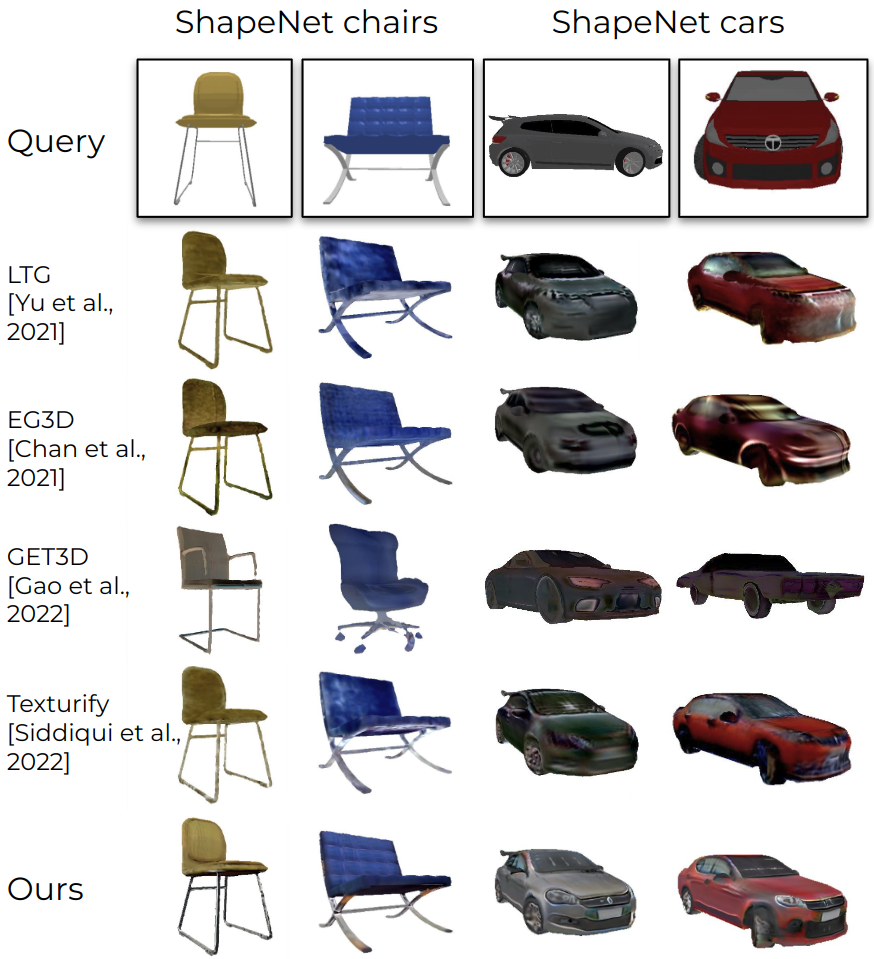}
    \vspace{-0.3cm}
    \caption{
    Texture transfer from synthetic image queries to exact shape geometry with unknown image alignment.
    \OURS{} produces more geometrically consistent texturing with finer-resolution details captured in our mesh-field texture representation. 
    Note that GET3D models both geometry and color, resulting in possible geometric changes.
    \vspace{-0.5cm}
    }
    \label{fig:opt_synth}
\end{center}%
\end{figure}

\paragraph{Exact geometry and known image alignment.}
We evaluate texture generation from a single image query to exactly matching object geometry with known image alignment in Tables~\ref{tab:aligned-comp} and \ref{tab:aligned-comp-clip-lpips}, which measure the distribution quality and perceptual match to the query image, respectively.
The optimization through our high-fidelity texture manifold enables more detailed texturing to match to various texture changes in the query image, in comparison to baselines.

\paragraph{Exact geometry and unknown image alignment.} 
We evaluate texture generation from a single image query where the pose of the shape in the input query is unknown. Quantitative evaluation with state of the art is shown in Tables~\ref{tab:non-aligned-comp} and \ref{tab:non-aligned-comp-clip-lpips}, which measure the distribution quality and perceptual match to the query image, respectively. 
Figure~\ref{fig:opt_synth} visualizes qualitative comparisons. 
For GET3D, we do not evaluate LPIPS, as it can produce geometric changes resulting in an inexact geometry match to the query.

Even with exact geometry, the unknown image alignment poses a notable challenge, resulting in degraded performance in the baselines. 
Our approach maintains more robustness due to the NOC guidance in our patch loss during texture optimization, producing more plausible texture generation.

\paragraph{Texture transfer.}
We evaluate texture transfer experiments from a single image query to an arbitrary shape of a different geometry than the input query. 
%
Tables~\ref{tab:out-of-dom-comp} and \ref{tab:out-of-dom-comp-clip-lpips} compare our approach to state of the art. 
This challenging scenario of differing geometry results in a performance drop in all methods; however, due to our NOC-guided style optimization on a mesh-field texture representation, \OURS{} can produce textures on novel shape geometry with greater coherency and finer details than baselines.

\subsubsection{Real Image Query Experiments}

\begin{figure}
\begin{center}
    \includegraphics[width=0.92\linewidth]{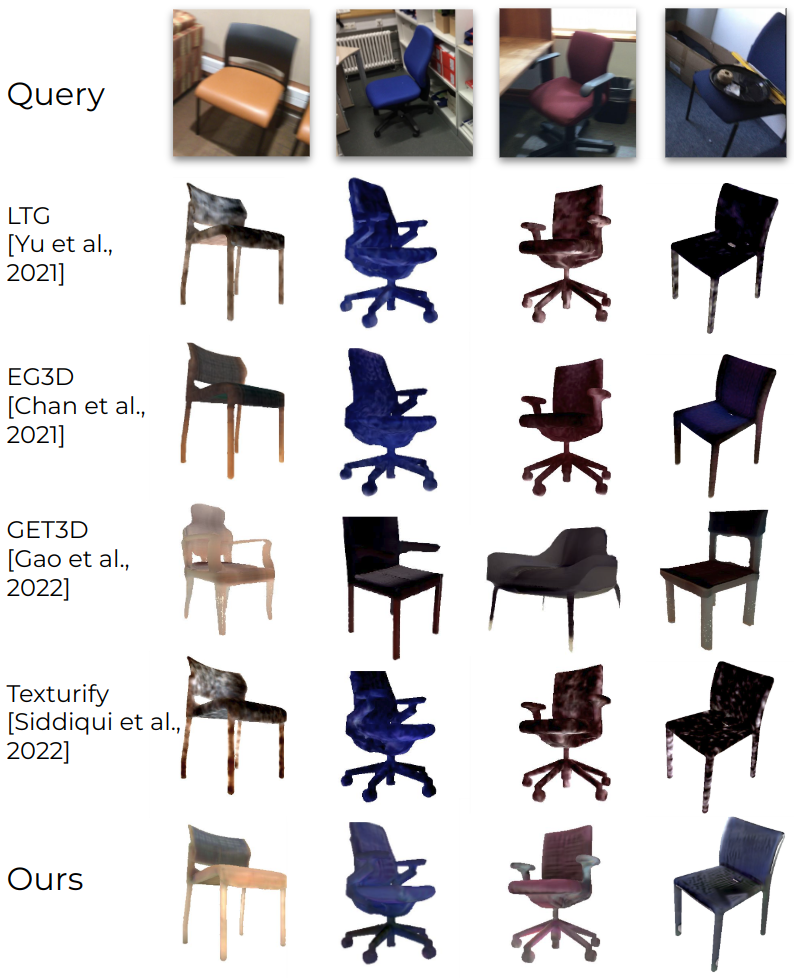}
    \vspace{-0.3cm}
    \caption{Texture transfer from real images (ScanNet) to closely matching shape geometry (ShapeNet).
    Despite inexact geometry and pose, \OURS{} produces perceptually consistent texturing.
    \vspace{-0.7cm}
    }
    \label{fig:opt_real}
\end{center}%
\end{figure}

\paragraph{Closely-matching geometry and image alignment.}
We evaluate texturing from a single real-world image, where the shape geometry to be textured and image alignment are close (as we cannot guarantee an exact match to real-world images).
%
We compare with state of the art in Tables~\ref{tab:aligned-comp} and \ref{tab:aligned-comp-clip-lpips}. 
In the real-world setting, view-dependent effects in real images (e.g., specular highlights, reflections) pose a notable challenge in evaluating texture optimization.
With close but inexact geometry and alignment, our method can still produce textures that are consistent with the image queries due to NOC-guided optimization, while baseline methods struggle to transfer textures from incomplete and cluttered real-world objects.

\paragraph{Closely-matching geometry and unknown image alignment.} 
We further consider texturing from a single real-world image to close shape geometry with unknown image alignment. 
We show qualitative results in Figure~\ref{fig:opt_real} and quantitative evaluation in Tables~\ref{tab:non-aligned-comp} and \ref{tab:non-aligned-comp-clip-lpips}. 
Our NOC-guided optimization through our texture manifold generates more perceptually representative textures.

\begin{figure}
\begin{center}
    \includegraphics[width=1.0\linewidth]{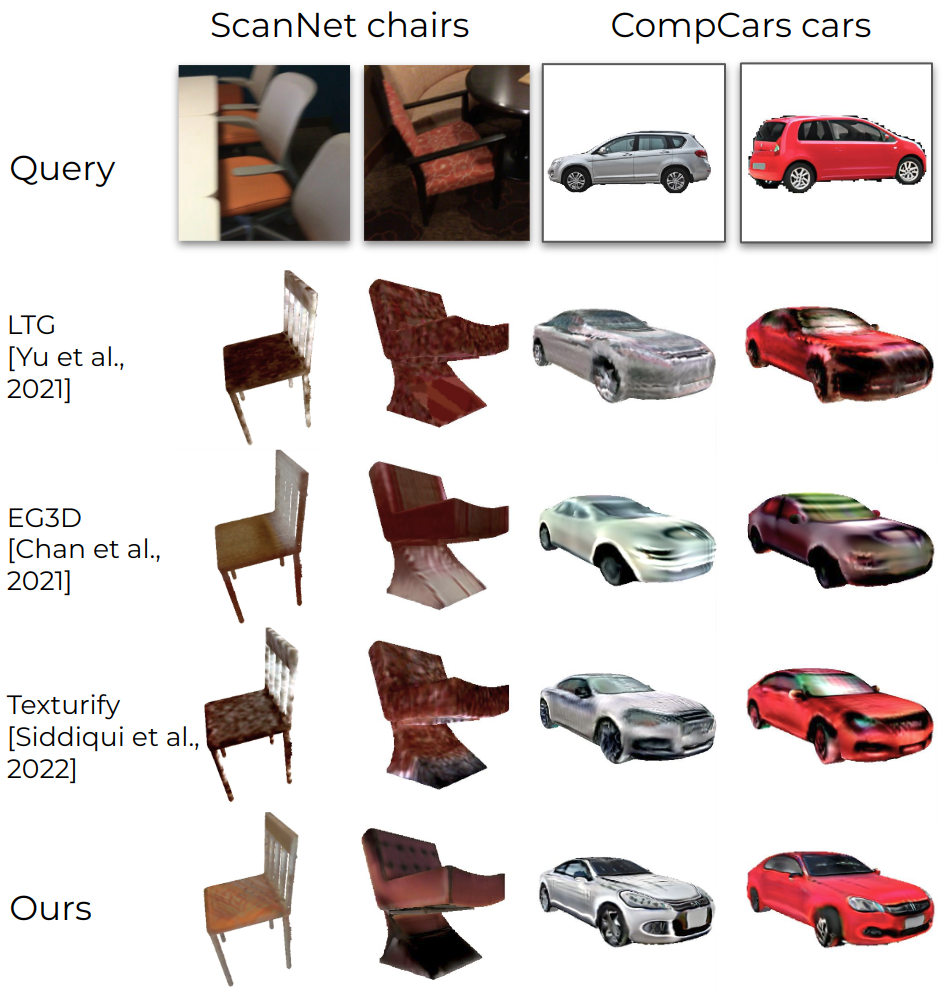}
    \vspace{-0.3cm}
    \caption{
    Texture transfer from real images (ScanNet, CompCars) to arbitrary shape geometry of the same class category (ShapeNet). Under this challenging scenario, our NOC-guided patch optimization enables plausible texturing for image queries.
    \vspace{-0.5cm}
    }
    \label{fig:opt_transfer}
\end{center}%
\end{figure}

\paragraph{Texture transfer.} 
Finally, we evaluate the challenging texture transfer scenario from a single real-world image from ScanNet~\cite{dai2017scannet} to an arbitrary ShapeNet~\cite{shapenet2015} mesh of the same class category.
This reflects potential texturing applications from users who wish to use an inspirational image to generate texturing on a given shape.
We show a quantitative comparison in Tables~\ref{tab:out-of-dom-comp} and \ref{tab:out-of-dom-comp-clip-lpips}, and qualitative results in Figure~\ref{fig:opt_transfer}. 
Our mesh-field texture representation maintains more detail with NOC-guided patch loss to provide more robustness during optimization.

\begin{table}[tp]
\centering
\resizebox{0.48\textwidth}{!}{
\begin{tabular}{l|cc|cc|cc}
\toprule
    \multirow{2}{*}{Method} & \multicolumn{2}{c|}{Chairs} & \multicolumn{2}{c|}{Cars} & \multicolumn{2}{c}{Chairs} \\
    & \multicolumn{2}{c|}{ShapeNet} & \multicolumn{2}{c|}{ShapeNet} & \multicolumn{2}{c}{ScanNet} \\
\cmidrule{2-7}
    & \multicolumn{4}{c|}{Exact geometry} & \multicolumn{2}{c}{Close geometry} \\
\cmidrule{2-7}
    & FID & KID & FID & KID & FID & KID \\
\midrule
    EG3D & 146.6 & 9.49 & 297.9 & 20.82 & 318.8 & 27.39 \\
    Yu et al. & 153.1 & 10.2 & 289.4 & 18.88 & 329.8 & 28.60 \\
    GET3D & 265.6 & 26.67 & 248.8 & 15.57 & 336.5 & 38.74 \\
    Texturify & 155.9 & 10.94 & 301.4 & 20.63 & 342.2 & 33.65 \\
\midrule
    {\bf Ours} & {\bf 115.4} & {\bf 5.38} & {\bf 165.5} & {\bf 6.84} & {\bf 309.7} & {\bf 26.50} \\
\bottomrule
\end{tabular}
}
\vspace{-0.2cm}
\caption{Evaluation on \textbf{\color{Fuchsia} aligned queried texture generation} with image queries from rendered ShapeNet chairs and cars, as well as real-world images from ScanNet chairs. \OURS{} generates more realistic textures from single-image queries. The KID values are scaled by $10^2$.
\vspace{-0.5cm}
}
\label{tab:aligned-comp}
\end{table}

\subsection{Ablations}

\noindent
\textbf{What is the impact of our mesh-field representation for texture generation?} 
Texturify~\cite{siddiqui2022texturify} represents a per-face mesh color generation approach, while we introduce a shared barycentric neural field to capture high-fidelity details.
Figure~\ref{fig:generation} shows our much finer-level detail generation, particularly for complex textures on cars.

\noindent
\textbf{What is the impact of patch style loss in the texture optimization process?} 
Table~\ref{tab:ablation} shows that our patch-based style loss significantly improves texture transfer optimization performance, allowing for capture of more local detail and robustness to mismatches to the query image.

\noindent
\textbf{What is the impact of NOC guidance for the patch style loss?}
In Table~\ref{tab:ablation}, we see that the NOC guidance helps to improve the perceptual quality of results by establishing coarse correspondence to the query image.


\noindent
\textbf{What is the effect or surface feature optimization?}
Following latent texture code optimization, we allow for surface features in the texture generator to further be optimized, which produces slightly improved local texture detail, as shown in Table~\ref{tab:ablation}.

\noindent
We refer to the supplemental for qualitative ablation analysis.

\begin{table}[tp]
\centering
\resizebox{0.48\textwidth}{!}{
\begin{tabular}{l|cc|cc|c}
\toprule
    \multirow{2}{*}{Method} & \multicolumn{2}{c|}{Chairs} & \multicolumn{2}{c|}{Cars} & \multicolumn{1}{c}{Chairs} \\
    & \multicolumn{2}{c|}{ShapeNet} & \multicolumn{2}{c|}{ShapeNet} & \multicolumn{1}{c}{ScanNet} \\
\cmidrule{2-6}
    & \multicolumn{4}{c|}{Exact geometry} & \multicolumn{1}{c}{Close geometry} \\
\cmidrule{2-6}
    & CLIP & LPIPS & CLIP & LPIPS & CLIP \\
\midrule
    EG3D & 89.87 & 16.41 & 83.27 & 15.69 & 82.86 \\
    Yu et al. & 89.73 & 16.26 & 84.33 & 14.83 & 82.93 \\
    GET3D & 85.00 & - & 81.38 & - & 78.52 \\
    Texturify & 89.22 & 16.23 & 83.82 & {\bf 14.69} & 81.71 \\
\midrule
    {\bf Ours} & {\bf 92.27} & {\bf 14.64} & {\bf 85.91} & 14.77 & {\bf 84.28} \\
\bottomrule
\end{tabular}
}
\vspace{-0.2cm}
\caption{Evaluation on \textbf{\color{Fuchsia} aligned queried texture generation} with image queries from rendered ShapeNet chairs and cars, as well as real-world images from ScanNet chairs.
Our texture transfer optimization produces more perceptually representative texturing results.
\vspace{-0.2cm}
}
\label{tab:aligned-comp-clip-lpips}
\end{table}

\begin{table}[tp]
\centering
\resizebox{0.48\textwidth}{!}{
\begin{tabular}{l|cc|cc|cc}
\toprule
    \multirow{2}{*}{Method} & \multicolumn{2}{c|}{Chairs} & \multicolumn{2}{c|}{Cars} & \multicolumn{2}{c}{Chairs} \\
    & \multicolumn{2}{c|}{ShapeNet} & \multicolumn{2}{c|}{ShapeNet} & \multicolumn{2}{c}{ScanNet} \\
\cmidrule{2-7}
    & \multicolumn{4}{c|}{Exact geometry} & \multicolumn{2}{c}{Close geometry} \\
\cmidrule{2-7}
    & FID & KID & FID & KID & FID & KID \\
\midrule
    EG3D & 155.2 & 9.13 & 326.8 & 24.48 & 319.4 & 27.18 \\
    Yu et al. & 160.0 & 9.88 & 343.3 & 32.71 & 327.1 & 28.20 \\
    GET3D & 271.9 & 26.54 & 254.4 & 16.63 & 337.3 & 38.76 \\
    Texturify & 174.2 & 11.74 & 344.2 & 25.34 & 333.3 & 28.81 \\
\midrule
    {\bf Ours} & {\bf 121.1} & {\bf 5.08} & {\bf 165.8} & {\bf 6.77} & {\bf 311.9} & {\bf 26.73} \\
\bottomrule
\end{tabular}
}
\vspace{-0.2cm}
\caption{Evaluation on \textbf{\color{NavyBlue} unaligned queried texture generation} with image queries from rendered ShapeNet chairs and cars, as well as real images from ScanNet chairs. 
\OURS{} maintains more realistic texture details than state of the art. The KID values are scaled by $10^2$.
\vspace{-0.2cm}
}
\label{tab:non-aligned-comp}
\end{table}

\begin{table}[tp]
\centering
\resizebox{0.48\textwidth}{!}{
\begin{tabular}{l|cc|cc|c}
\toprule
    \multirow{2}{*}{Method} & \multicolumn{2}{c|}{Chairs} & \multicolumn{2}{c|}{Cars} & \multicolumn{1}{c}{Chairs} \\
    & \multicolumn{2}{c|}{ShapeNet} & \multicolumn{2}{c|}{ShapeNet} & \multicolumn{1}{c}{ScanNet} \\
\cmidrule{2-6}
    & \multicolumn{4}{c|}{Exact geometry} & \multicolumn{1}{c}{Close geometry} \\
\cmidrule{2-6}
    & CLIP & LPIPS & CLIP & LPIPS & CLIP \\
\midrule
    EG3D & 89.51 & 16.83 & 81.93 & 16.14 & 83.40 \\
    Yu et al. & 88.98 & 16.81 & 79.35 & 21.84 & 82.67 \\
    GET3D & 84.66 & - & 81.54 & - & 79.02 \\
    Texturify & 88.24 & 16.82 & 80.45 & 15.42 & 82.26 \\
\midrule
    {\bf Ours} & {\bf 91.57} & {\bf 14.96} & {\bf 85.42} & {\bf 14.97} & {\bf 84.28} \\
\bottomrule
\end{tabular}
}
\vspace{-0.2cm}
\caption{Evaluation on \textbf{\color{NavyBlue} unaligned queried texture generation} to ShapeNet chairs and cars from rendered ShapeNet objects as well as real-world image queries from ScanNet chairs. 
Our NOC-guided patch style loss maintains more robustness to the unknown image alignment.
\vspace{-0.2cm}
}
\label{tab:non-aligned-comp-clip-lpips}
\end{table}

\begin{table}[tp]
\centering
\resizebox{0.48\textwidth}{!}{
\begin{tabular}{l|cc|cc|cc|cc}
\toprule
    \multirow{2}{*}{Method} & \multicolumn{2}{c|}{Chairs} & \multicolumn{2}{c|}{Cars} & \multicolumn{2}{c|}{Chairs} & \multicolumn{2}{c}{Cars} \\
    & \multicolumn{2}{c|}{ShapeNet} & \multicolumn{2}{c|}{ShapeNet} & \multicolumn{2}{c|}{ScanNet} & \multicolumn{2}{c}{CompCars} \\
\cmidrule{2-9}
    & FID & KID & FID & KID & FID & KID & FID & KID \\
\midrule
    EG3D & 297.4 & 28.53 & 343.6 & 26.57 & 355.2 & 33.31 & 352.3 & 39.15 \\
    Yu et al. & 303.4 & 29.10 & 351.1 & 26.75 & 362.6 & 34.20 & 385.4 & 51.37 \\
    GET3D & - & - & - & - & - & - & - & - \\
    Texturify & 307.9 & 29.63 & 357.1 & 27.05 & 370.4 & 34.85 & 264.7 & 29.23 \\
\midrule
    {\bf Ours} & {\bf 276.7} & {\bf 26.17} & {\bf 194.5} & {\bf 10.11} & {\bf 348.5} & {\bf 32.33} & {\bf 203.7} & {\bf 21.48} \\
\bottomrule
\end{tabular}
}
\vspace{-0.2cm}
\caption{Evaluation on \textbf{\color{BrickRed} texture transfer} to ShapeNet chairs and cars from rendered ShapeNet objects as well as real-world image queries from ScanNet chairs and CompCars cars. 
In this challenging scenario, our mesh-field representation retains more realistic detail in texture optimization. The KID values are scaled by $10^2$.
\vspace{-0.3cm}
}
\label{tab:out-of-dom-comp}
\end{table}

\begin{table}[tp]
\centering
\resizebox{0.48\textwidth}{!}{
\begin{tabular}{l|c|c|c|c}
\toprule
    \multirow{2}{*}{Method} & \multicolumn{1}{c|}{Chairs} & \multicolumn{1}{c|}{Cars} & \multicolumn{1}{c|}{Chairs} & \multicolumn{1}{c}{Cars} \\
    & \multicolumn{1}{c|}{ShapeNet} & \multicolumn{1}{c|}{ShapeNet} & \multicolumn{1}{c|}{ScanNet} & \multicolumn{1}{c}{CompCars} \\
\cmidrule{2-5}
    & CLIP & CLIP & CLIP & CLIP\\
\midrule
    EG3D & 81.62 & 80.72 & 81.54 & 68.18 \\
    Yu et al. & 81.31 & 78.69 & 81.21 & 64.76 \\
    GET3D & - & - & - & - \\
    Texturify & 80.70 & 78.59 & 80.99 & 66.93 \\
\midrule
    {\bf Ours} & {\bf 83.47} & {\bf 82.20} & {\bf 81.95} & {\bf 77.71} \\
\bottomrule
\end{tabular}
}
\vspace{-0.2cm}
\caption{Evaluation on \textbf{\color{BrickRed} texture transfer} to ShapeNet chairs and cars from rendered ShapeNet objects as well as real-world image queries from ScanNet chairs and CompCars cars. 
Our NOC-guided patch style loss enables better characterization of the input query image for texturing.
\vspace{-0.3cm}
}
\label{tab:out-of-dom-comp-clip-lpips}
\end{table}

\begin{table}[tp]
\centering
\resizebox{0.37\textwidth}{!}{
\begin{tabular}{l|ccc}
\toprule
    \multirow{2}{*}{Method} & \multicolumn{3}{c}{Chairs (unaligned)} \\
    & \multicolumn{3}{c}{ShapeNet} \\
\cmidrule{2-4}
    & FID & KID & CLIP \\
\midrule
    w/o NOC & 124.3 & 5.53 & 91.09 \\
    w/o patches & 143.5 & 6.91 & 89.68 \\
    w/o surf. features & 127.1 & 5.80 & 90.70 \\
\midrule
    {\bf Ours} & {\bf 121.1} & {\bf 5.08} & {\bf 91.57} \\
\bottomrule
\end{tabular}
}
\vspace{-0.2cm}
\caption{\textbf{Ablation study} on texture transfer optimization design.
Our NOC guidance, patch style loss, and surface feature optimization help to improve generated texture quality. The KID values are scaled by $10^2$.
\vspace{-0.4cm}
}
\label{tab:ablation}
\end{table}

\paragraph{Limitations.}
\OURS{} offers a promising step towards conditional mesh texturing from a single image, though various limitations remain.
For instance, our approach does not explicitly model semantics, potentially leading to distortions in texture in semantically meaningful areas (e.g., spokes of a car wheel).
Additionally, a more explicit characterization of the texture distribution could enable probabilistic sampling when input queries may be occluded or incomplete.
\section{Conclusion}

We presented \OURS{}, which learns a texture manifold based on a hybrid mesh-field representation to generate realistic, high-quality textures on shape geometry from real-world imagery.
\OURS{} enables test-time optimization for texture transfer from single image queries.
Crucially, our NOC-guided local and global style loss enables optimization for perceptually matching textures on a 3D shape that does not require exact geometric matches to RGB image queries.
We believe this opens up many new possibilities in texturing for content creation, enabling easy-to-capture images to be used as guidance for holistic shape texturing.

\medskip
\noindent \textbf{Acknowledgements.}
This work was supported by the Bavarian State Ministry of Science and the Arts coordinated by the Bavarian Research Institute for Digital Transformation (bidt). ST was partially supported by a gift award from CISCO.

{\small
\bibliographystyle{ieee_fullname}
\bibliography{egbib}
}

\clearpage
\appendix

In this supplemental material, we show additional texture generation results in Section~\ref{sec:supp_gen}, texture transfer from query images in Section~\ref{sec:supp_texturing}, qualitative visualizations for ablations in Section~\ref{sec:supp_ablation}, network architecture details in Section~\ref{sec:supp_arch}, and additional implementation details in Section~\ref{sec:supp_impl}.

\begin{figure*}[bp]
\begin{center}
    \includegraphics[width=0.98\linewidth]{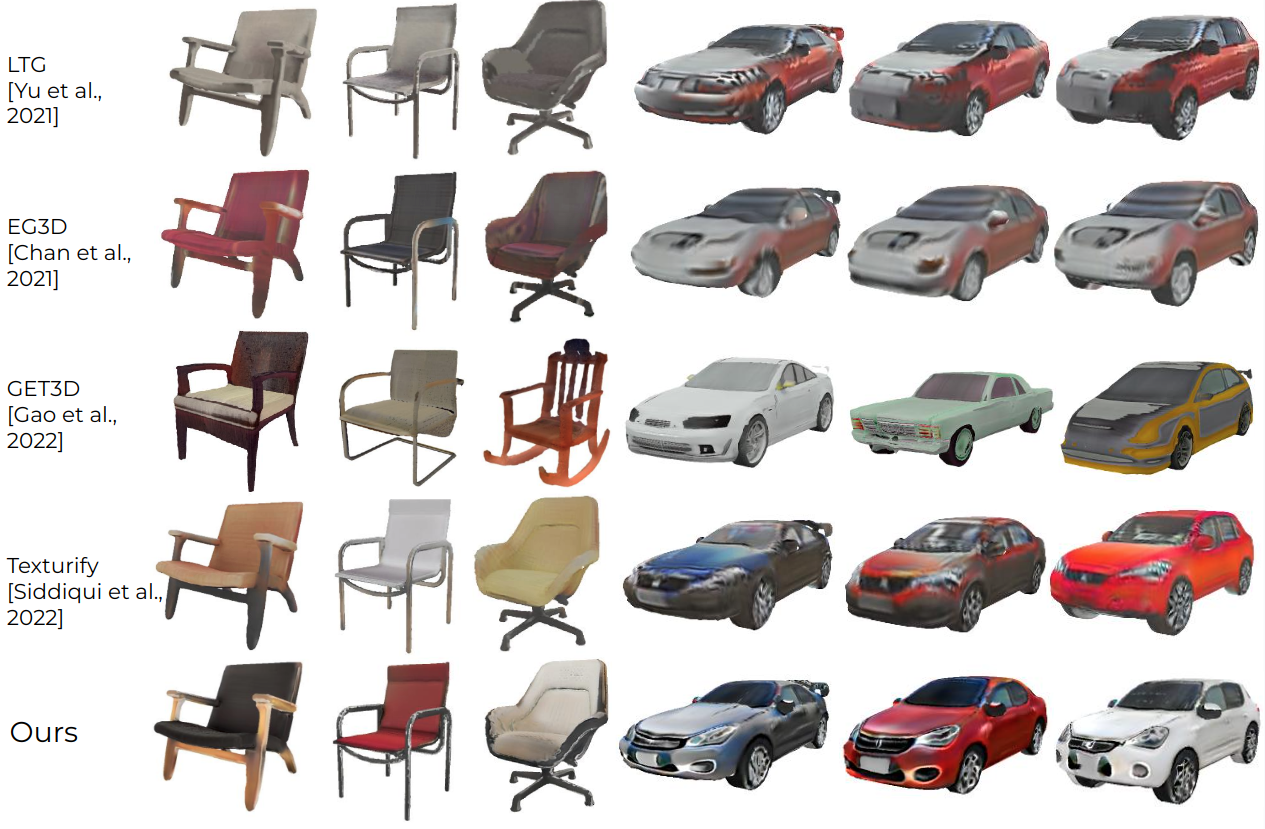}
    \vspace{-0.4cm}
    \caption{
    Additional results on unconditional texturing for meshes from ShapeNet~\cite{shapenet2015}, in comparison with state of the art.
    Our approach generates more realistic, detailed textures.
    }
    \label{fig:supp_generation}
\end{center}%
\end{figure*}

\section{Additional Generation Results}
\label{sec:supp_gen}

In Fig.~\ref{fig:supp_generation} we show additional qualitative evaluation on unconditional texture generation for ShapeNet meshes. Not all baselines are able to produce diverse high-quality textures after training on uncorrelated real-world datasets. As GET3D cannot be successfully trained on sparse views, we use the dense view per-object training provided by the authors; however, this still produces  different artifacts. Our learned hybrid mesh-field  representation enables \OURS{} to generate more realistic textures.

\section{Additional Results for Texture Transfer from a Single RGB Image}
\label{sec:supp_texturing}

\begin{figure*}[tp]
\begin{center}
    \includegraphics[width=0.98\linewidth]{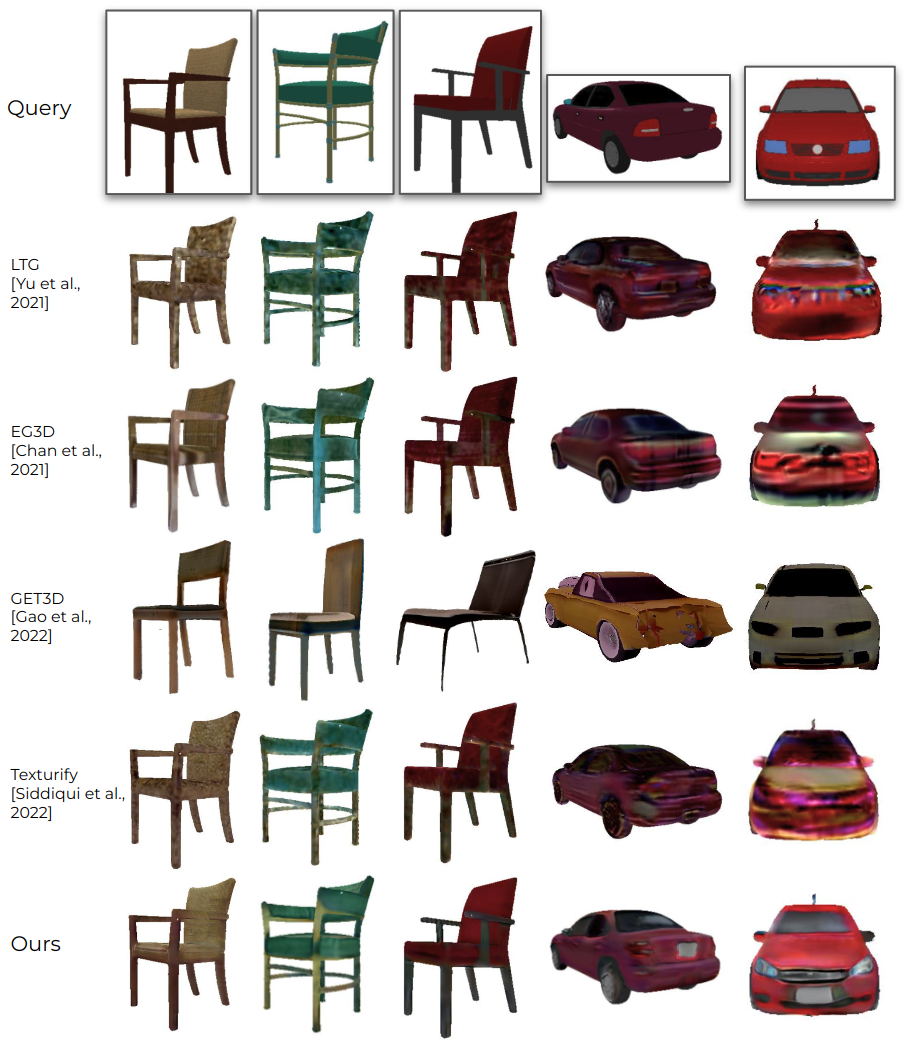}
    \vspace{-0.4cm}
    \caption{
    Optimized textures based on input query images (top row) using aligned query images from ShapeNet chairs and cars.
    \vspace{1.0cm}
    }
    \label{fig:supp_synth_aligned}
\end{center}%
\end{figure*}

\begin{figure*}
\begin{center}
    \includegraphics[width=0.98\linewidth]{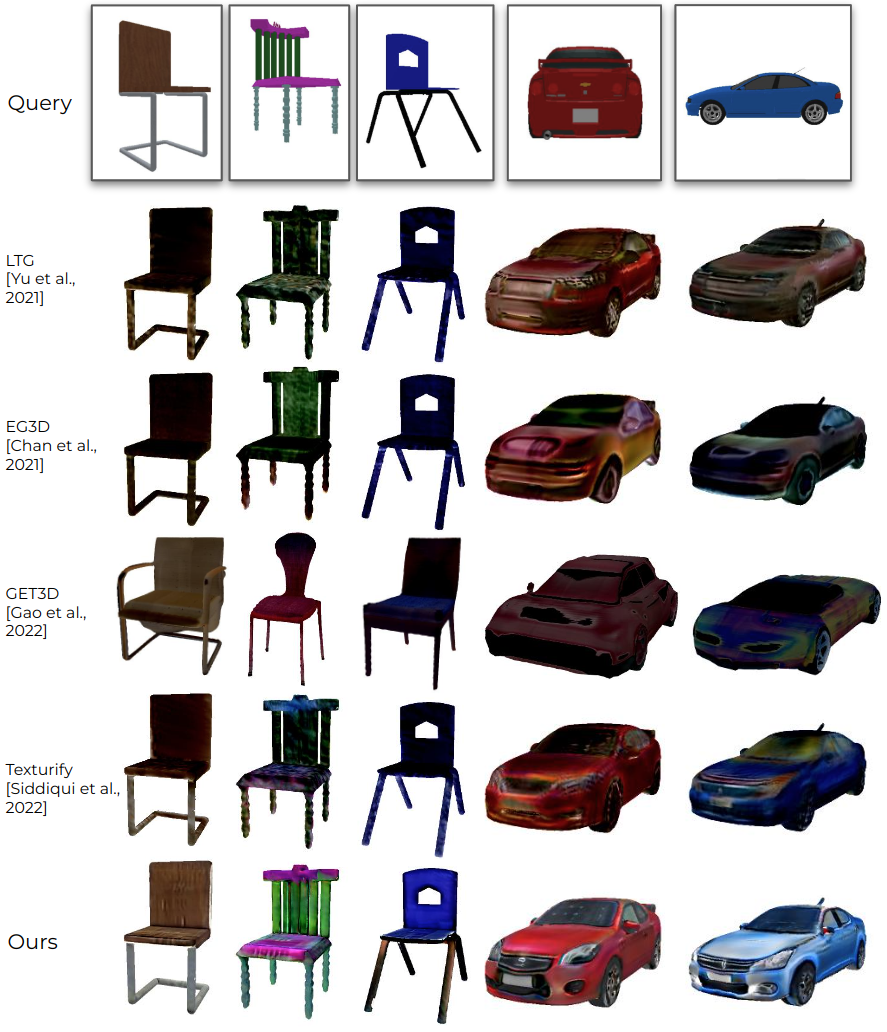}
    \vspace{-0.4cm}
    \caption{
   Optimized textures based on input query images (top row) using unaligned images  query images from ShapeNet chairs and cars.
    \vspace{1.0cm}
    }
    \label{fig:supp_synth_unaligned}
\end{center}%
\end{figure*}

\begin{figure*}
\begin{center}
    \includegraphics[width=0.98\linewidth]{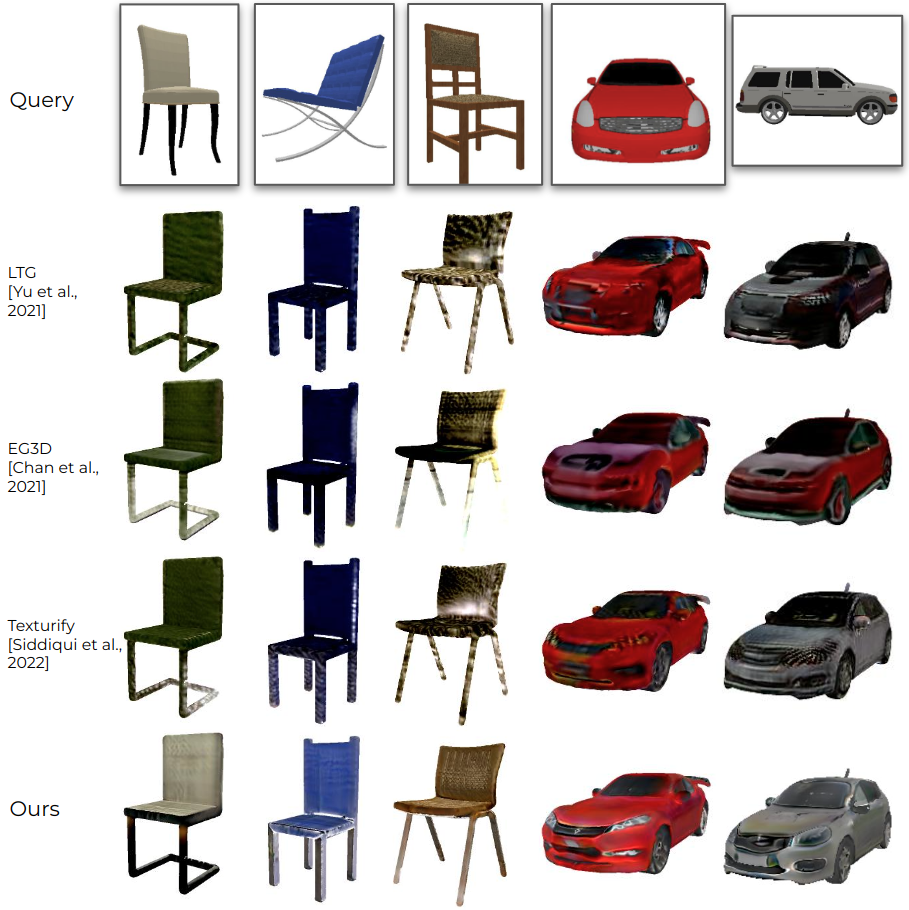}
    \vspace{-0.4cm}
    \caption{
    Texture transfer from input query images (top row) using unaligned images and arbitrary shape geometry of the same class category (ShapeNet).
    }
    \label{fig:supp_synth_transfer}
\end{center}%
\end{figure*}

In Figs.~\ref{fig:supp_synth_aligned}, \ref{fig:supp_synth_unaligned}, and \ref{fig:supp_synth_transfer}, we present additional results on for texture transfer from synthetic input images, using aligned, unaligned query images and arbitrary shape geometry, respectively. \OURS{} effectively leverages our learned texture manifold, preserving consistent textures while transferring to different geometries.

\begin{figure*}
\begin{center}
    \includegraphics[width=0.91\linewidth]{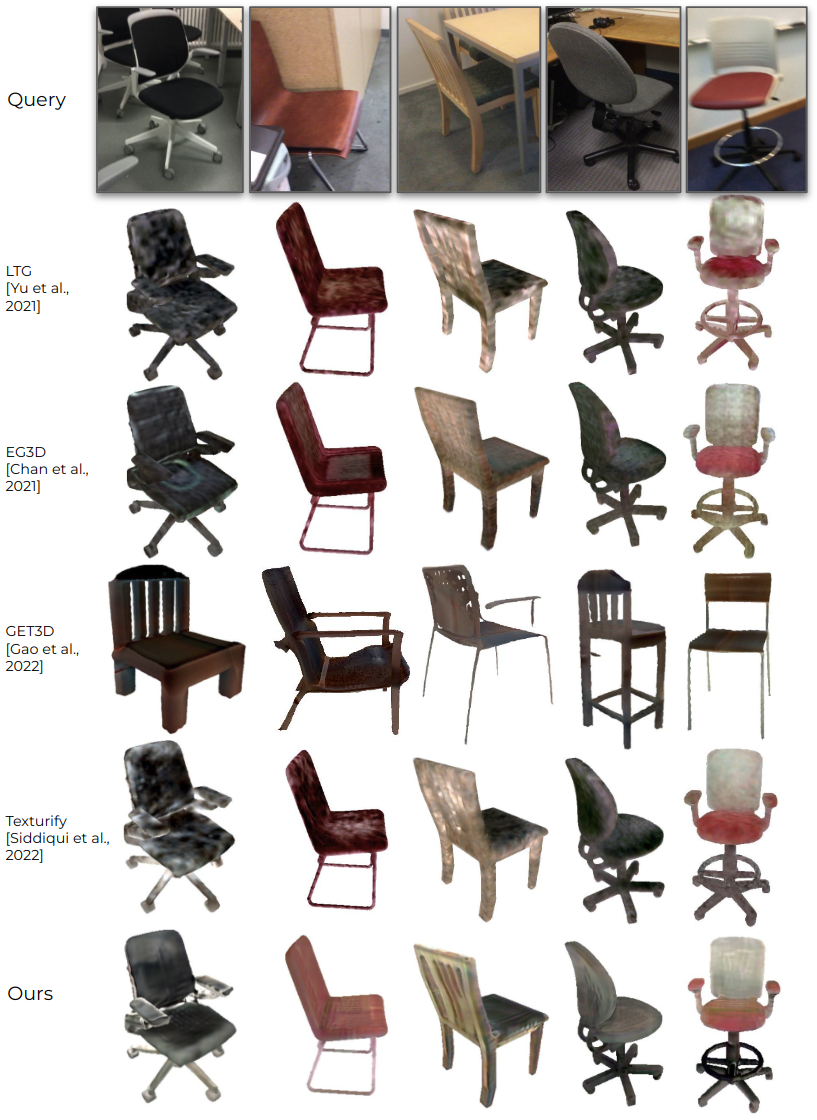}
    \vspace{-0.4cm}
    \caption{
    Texture transfer from real-world input query images (top row, ScanNet) using aligned images and close shape geometry (ShapeNet).
    }
    \label{fig:supp_real_aligned}
\end{center}%
\end{figure*}

\begin{figure*}
\begin{center}
    \includegraphics[width=0.91\linewidth]{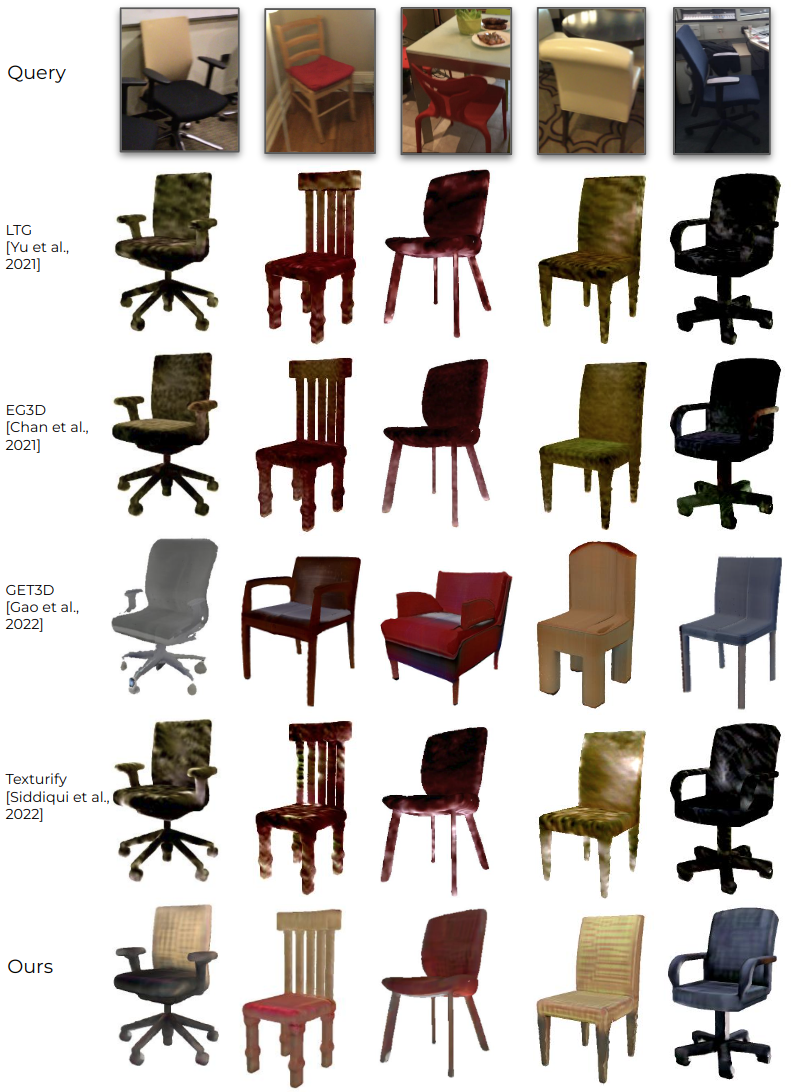}
    \vspace{-0.4cm}
    \caption{
    Texture transfer from real-world input query images (top row, ScanNet) using unaligned images and similar shape geometry (ShapeNet).
    }
    \label{fig:supp_real_unaligned}
\end{center}%
\end{figure*}

\begin{figure*}
\begin{center}
    \includegraphics[width=0.91\linewidth]{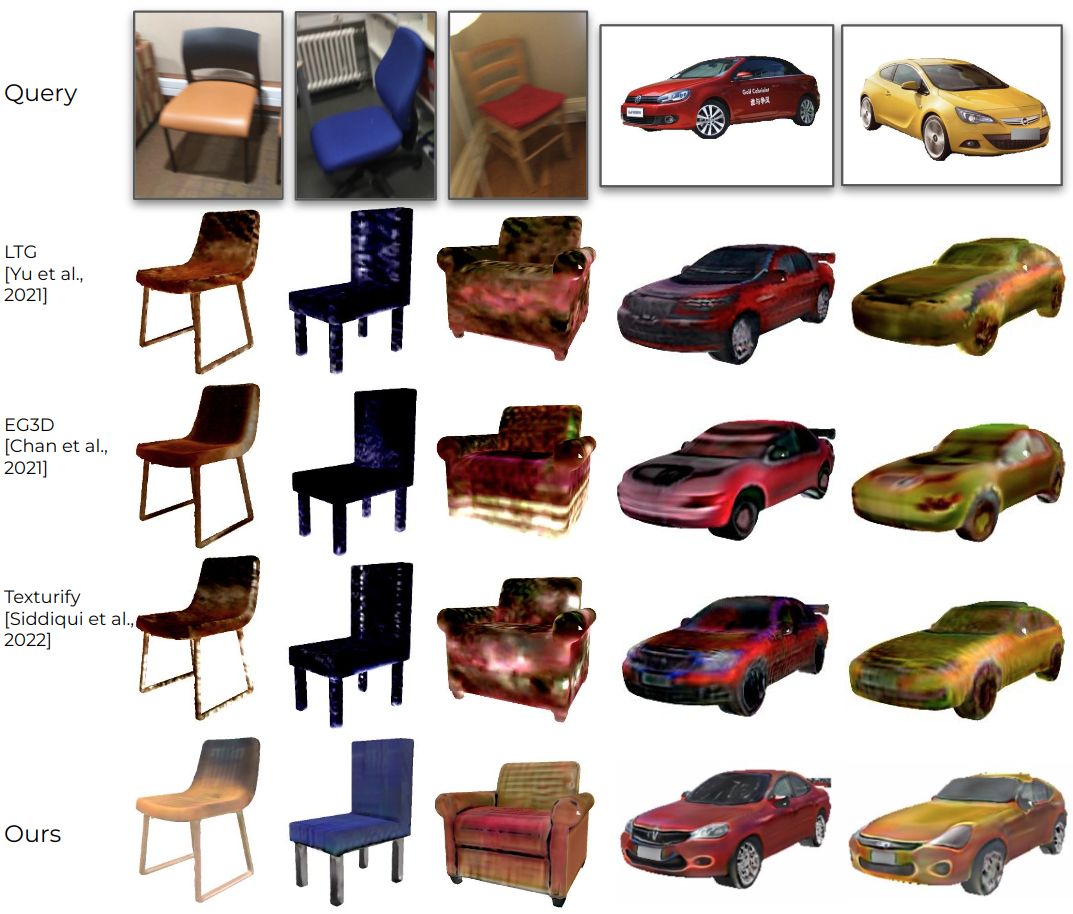}
    \vspace{-0.4cm}
    \caption{
    Texture transfer from real-world input query images (top row, ScanNet, CompCars) using unaligned images and arbitrary shape geometry from the same class category (ShapeNet).
    }
    \label{fig:supp_real_transfer}
\end{center}%
\end{figure*}

Figs.~\ref{fig:supp_real_aligned}, \ref{fig:supp_real_unaligned}, and \ref{fig:supp_real_transfer} present additional results on texture transfer from  real-world images from ScanNet and CompCars images as queries. \OURS{} is able to perform consistent texture generation from real-world queries even under the challenging scenario of different geometry, pose, and real-world view-dependent effects.

\begin{figure*}
\begin{center}
    \includegraphics[width=0.93\linewidth]{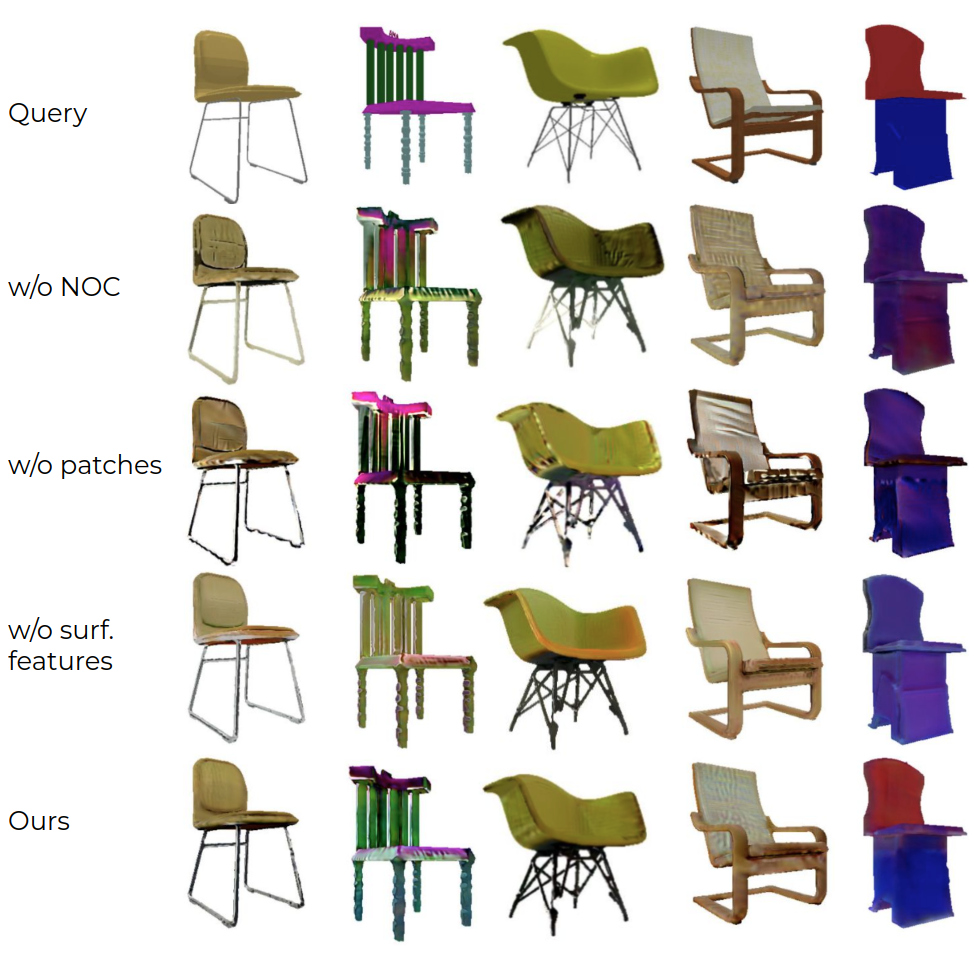}
    \vspace{-0.4cm}
    \caption{
    Qualitative ablation study on texture transfer from synthetic input query images (ShapeNet) using unaligned images; visualized with the query image pose.
    }
    \label{fig:supp_synth_ablation}
\end{center}%
\end{figure*}

\section{Qualitative Ablation Visualizations}
\label{sec:supp_ablation}

According to Tab.8 in the main paper, we show qualitative results on an ablation study performed in texture generation from unaligned synthetic ShapeNet images. Optimizing textures without patch loss component or NOC guidance leads to messy textures with stripe artifacts and disordered texture mapping. Optimizing only the latent codes (w/o surface features) results in inaccurate texture generation with lost details.

\section{Network Architecture}
\label{sec:supp_arch}

\begin{figure*}
\begin{center}
    \includegraphics[width=1.0\textwidth]{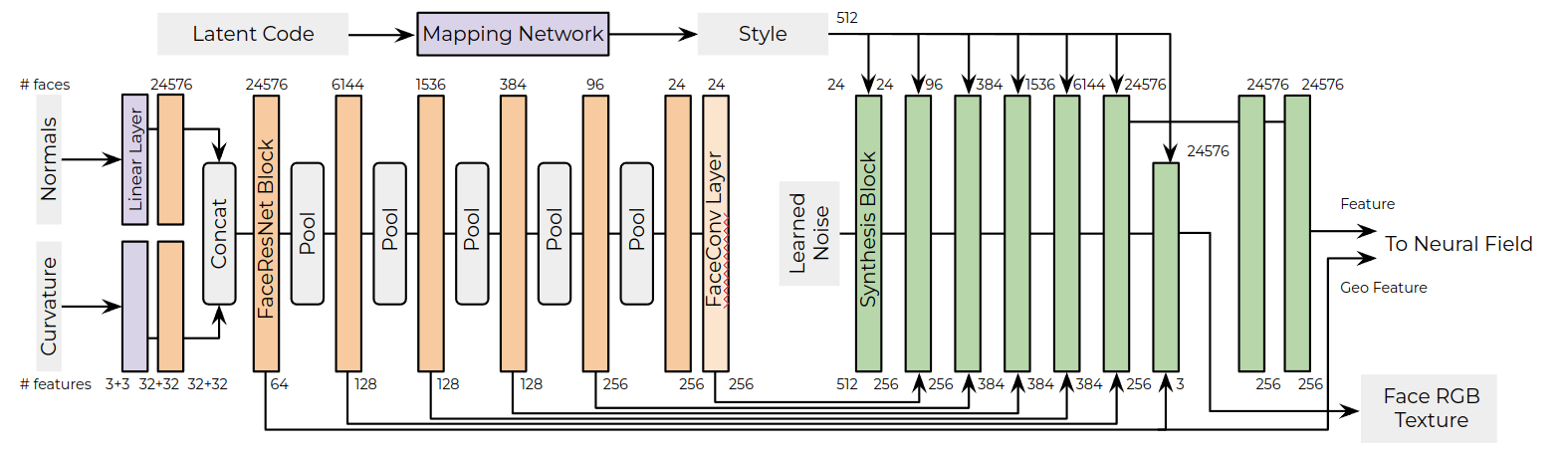}
    \caption{Overview of Surface Encoder (orange) and Generator (green) architectures. The encoder takes as input normals and curvatures of the finest resolution quadmesh, processes them in a hierarchical convolutional structure to extract geometric features. The generator then considers a latent texture code, learned noise, and the geometric features to produce per-face features for the neural field. 
    }
    \label{fig:main_arch}
\end{center}%
\end{figure*}

\begin{figure*}
\begin{center}
    \includegraphics[width=1.0\textwidth]{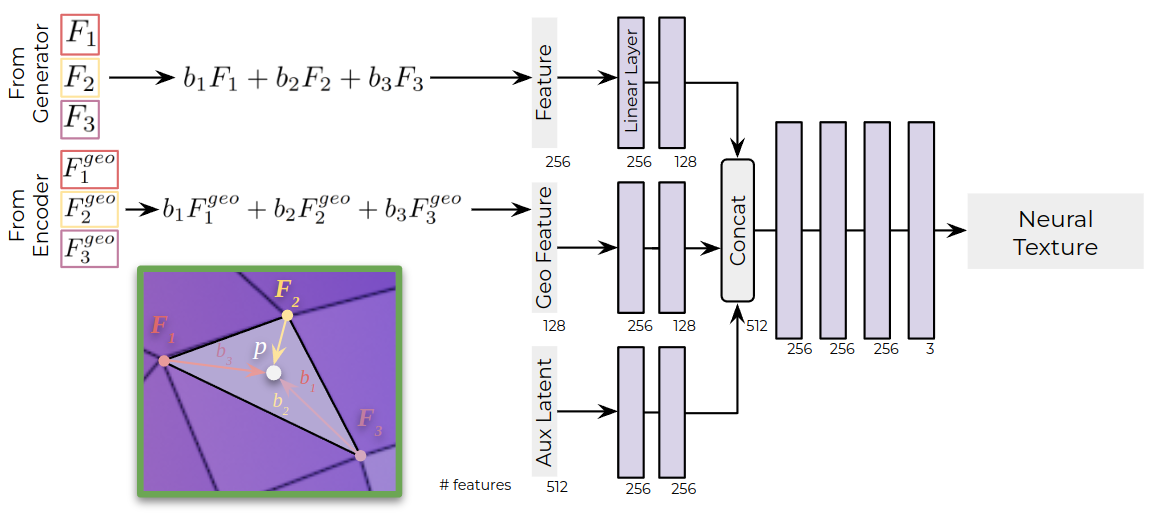}
    \caption{Our architecture for local face neural field $\Psi$.  
    Features from the surface encoder and generator are fused by their barycentric coordinates and passed with an auxiliary latent vector into an MLP to produce the final color of surface location $p$.
    }
    \label{fig:renderer_arch}
\end{center}%
\end{figure*}

An overview of the architectures of our surface encoder, generator, and neural field $\Psi$  is shown in Figs.~\ref{fig:main_arch} and \ref{fig:renderer_arch}. Note that we follow the use of FaceResNet blocks, FaceConv layers, and Synthesis Block from Texturify~\cite{siddiqui2022texturify}. 
Our generator then produces both coarse per-face rgb values as well as feature vector $F_c$ input to $\Psi$.
Our discriminator architecture follows the discriminator of Texturify.

In order to produce locally refined texture with $\Psi$, we operate locally on faces, considering their barycentric coordinate system. 
We compute per-vertex features from $F_c$ by averaging incident face features. For a point $p$ on the mesh surface, its feature is computed as the barycentric averaging of the vertex features $F_1, F_2, F_3$, where the barycentric weights $b_1, b_2, b_3$ are areas of triangles $\vartriangle~(F_1, F_2, p), \vartriangle~(F_2, F_3, p), \vartriangle~(F_3, F_1, p)$ respectively. 
$\Psi$ also takes a learnable auxiliary latent vector of size $z_{aux}=512$ as input, which is fixed for the entire model.
We observed that this auxiliary latent enhanced the consistency of high-resolution textures. The auxiliary latent, along with surface encoder and generator features, are then processed with two linear layers (LeakyReLU activations) before being concatenated and passed to an additional four linear layers. This results in the final output color corresponding to surface location  $p$.

When optimizing for texture from an input image query, we also use a pose predictor network and NOC predictor network.
Pose prediction uses a ResNet18~\cite{he2016deep} backbone, with the final features of size $512$ passed to two linear layers with output dimension $256$. This refined feature is then passed to a two-layer MLP with hidden size $128$ to estimate the angles $\alpha^a$ and $\alpha^e$. 
The NOC predictor leverages the EfficientNet-b4~\cite{Tan2019EfficientNetRM} architecture as a backbone for the U-shaped model. It takes an RGB image and the corresponding binary mask of a foreground object as 4-channel input and predicts NOCs a 3-channel image.

\section{Implementation Details}
\label{sec:supp_impl}

We train Mesh2Tex using an Adam optimizer with learning rates of 1e-4, 12e-4, 1e-4, 14e-4 for the encoder, generator, $\Psi$, and both discriminators, respectively, for PhotoShape~\cite{photoshape2018}, and learning rates of 1e-4, 15e-4, 5e-4, 1e-4 for CompCars~\cite{yang2015large}. 
For both models, we use a batch size of 2, and render 8 views for each shape in the batch.

To optimize texturing for input query images, we optimize latent codes for 100 iterations and refined weights with the parameters of the two last generator synthesis blocks for additional 300 iterations using an Adam optimizer with a learning rate of 1e-2. For the style loss, we use VGG19~\cite{SimonyanZ14a} network, and extract features of the $(2^{nd}, 4^{th}, 8^{th}, 12^{th}, 16^{th})$ convolutional layers. We extract patches of size 64 from both rendered and input images. Both full images and patches are equipped with corresponding foreground masks to filter out extracted VGG background features. 

Pose prediction is trained using an Adam optimizer with a learning rate of 3e-4 for chairs and 2e-5 for cars, with a batch size of 128 on images of size 512x512. The NOC Predictor is trained using an Adam optimizer with a learning rate of 3e-4 for chairs and 5e-5 for cars, with a batch size of 64 on images of size 512x512. Both networks are first pretrained on synthetic renders of ShapeNet objects and fine-tuned on real-world ScanNet and CompCars datasets (except for NOC Prediction for CompCars as NOC data is not available).


\end{document}